\documentclass{article}

\usepackage{neurips_2026}
\nipsfinalcopy

\usepackage[utf8]{inputenc}
\usepackage[T1]{fontenc}
\usepackage{hyperref}
\usepackage{url}
\usepackage{booktabs}
\usepackage{amsfonts}
\usepackage{amsmath}
\usepackage{nicefrac}
\usepackage{microtype}
\usepackage{graphicx}
\usepackage{xcolor}
\usepackage{float}
\usepackage{subcaption}
\usepackage{multirow}
\usepackage{array}
\usepackage{tikz}
\usepackage{pgfplots}
\usetikzlibrary{arrows.meta,positioning,calc,fit,backgrounds}
\pgfplotsset{compat=1.18}
\newcommand{\model}{\textsc{ViralityNet}}
\newcommand{\eg}{\emph{e.g.}}

\title{Predicting Post Virality with Temporal Cross-Attention 
       over Trend Signals}

\author{
\begin{tabular}{cc}
Sarvagya Somvanshi & Mohan Xu \\
\texttt{sarvagya@berkeley.edu} & \texttt{mohanxu@berkeley.edu} \\
\\
Rakhi Chadalavada & Nathan Canera \\
\texttt{rakhichd@berkeley.edu} & \texttt{nathancanera@berkeley.edu}
\end{tabular}
}

\begin{document}

\maketitle

% ── Abstract ─────────────────────────────────────────────────────────────────
\begin{abstract}
Current models for predicting social media virality rely heavily on static textual and structural features, effectively ignoring the highly dynamic nature of trend signals.
We study whether real-world attention signals can improve the prediction of social-media virality beyond what post text alone reveals. We introduce \model{}, an architecture that predicts Reddit post virality by fusing internal platform representations with exogenous temporal signals derived from Wikipedia pageview spikes. We frame virality as a binary classification task that accounts for differences in subreddit scale, labeling posts as viral if they exceed the 90th percentile of per-subreddit engagement and a minimum absolute score threshold.

\model{} combines four post-level streams---title embeddings, body embeddings, structural metadata, and learned subreddit embeddings---with a cross-attention block that queries a daily sliding-window trends matrix encoding the top-512 Wikipedia spike terms from the preceding seven days. Empirical results suggest that incorporating external attention signals yields consistent gains, outperforming text-only baselines by +0.015 AUC-PR and achieving an overall AUC-ROC of 0.836. Overall, we provide evidence that incorporating external attention signals yields measurable improvements over text-only baselines, highlighting the importance of real-world dynamics in shaping online virality.

\end{abstract}

% ── 1. Introduction ──────────────────────────────────────────────────────────
\section{Introduction}

Understanding what makes online content go viral has attracted sustained
research attention from human-computer interaction, computational social
science, and natural language processing
communities~\cite{cheng2014can,szabo2010predicting}. Most prior work has
focused on structural features including follower counts, posting time, network
centrality, and textual semantics~\cite{tan2014effect}. 

However, a key limitation of purely text-based models is that they are
\emph{temporally blind}: a post about a topic that was intensely
discussed last month may carry very different viral potential from an
otherwise identical post written during a week when that topic is spiking
in public attention.
This limitation matters because virality is not determined by content
alone, but also by how well that content aligns with what the public
cares about at a given moment.
Models that ignore these temporal dynamics risk mischaracterizing
the drivers of engagement, limiting both predictive accuracy and
scientific understanding of online behaviour.

We hypothesise that internet-wide attention signals, measured via daily
Wikipedia pageview spikes, serve as a proxy for collective public
interest that can inform virality prediction on Reddit.
When a topic is rapidly gaining attention (reflected in a sudden surge
of Wikipedia reads), Reddit posts about that topic are more likely to
find a receptive audience.

\paragraph{Contributions.}
We make three contributions:
\begingroup
\setlength{\itemsep}{2pt}
\setlength{\parskip}{0pt}
\setlength{\parsep}{0pt}
\begin{itemize}
  \item We introduce a post-level trend alignment signal derived from Wikipedia pageview spikes and show it is statistically significant at the individual post level.
  \item We propose \model{}, a cross-attention architecture that conditions post representations on daily sliding-window trend matrices, together with learned year and subreddit embeddings.
  \item We conduct a systematic ablation study that disentangles the contributions of temporal context (year embeddings), exogenous trends (cross-attention), and their combination.
\end{itemize}
\endgroup

{\setlength{\abovedisplayskip}{2pt}
 \setlength{\belowdisplayskip}{2pt}}
\paragraph{Virality score.}
We define a per-subreddit engagement score that combines upvotes and
comments:
{\setlength{\abovedisplayshortskip}{2pt}
 \setlength{\belowdisplayshortskip}{2pt}}
\begin{equation}
  v = s + \beta \cdot c, \quad \beta = 0.3,
  \label{eq:virality}
\end{equation}
where $s$ is the post score (upvotes minus downvotes) and $c$ is the
comment count.
The comment weight $\beta = 0.3$ reflects that comments represent greater
engagement effort than passive voting; a sensitivity analysis
(Appendix~\ref{app:sensitivity}) confirms that varying
$\beta \in [0.1, 0.5]$ flips fewer than 0.2\% of labels. 

\paragraph{Binary labelling.}
We adopt a \textbf{subreddit-stratified top-10\% labelling strategy}: within
each subreddit, a post is labelled viral ($y=1$) if $v$ exceeds the 90th
percentile of that subreddit's score distribution \emph{and} the raw score
$s \geq 100$. All posts falling below this threshold are strictly retained in the dataset as negative samples ($y=0$). The absolute-score floor prevents posts from tiny subreddits from being labelled viral despite negligible real-world engagement, while retaining the overwhelming majority of posts as negative examples provides the necessary class imbalance to train a robust discriminative model. Per-subreddit stratification is important because engagement scales differ by orders of magnitude across communities.

% ── 2. Related Work ───────────────────────────────────────────────────────────
\section{Related Work}
\label{sec:related}

\paragraph{Virality and popularity prediction.}
Predicting which content will achieve outsized engagement has been studied
across platforms and modalities.
Szabo and Huberman~\cite{szabo2010predicting} demonstrated a log-linear
relationship between early view counts and eventual popularity, showing
that the first hours of engagement strongly predict final reach.
Cheng et al.~\cite{cheng2014can} extended this to information cascades,
finding that cascade structure carries predictive signal beyond simple
popularity counts.
Tatar et al.~\cite{tatar2014predicting} found that news article engagement follows a heavy-tailed distribution, with the top 10\% of articles on major European platforms attracting roughly half of all comments. This skewed pattern both justifies treating virality as a binary threshold and demonstrates that even simple feature sets can yield meaningful predictions.
On Reddit specifically, Lakkaraju et al.~\cite{lakkaraju2013whats}
studied the role of title phrasing and submission timing, while Hessel et al.~\cite{hessel2017cats} used multimodal content together with community and temporal features to predict post popularity.
Most of these approaches treat posts in isolation from broader societal
context; our work differs by conditioning predictions on exogenous signals
of real-world public attention.

\paragraph{Exogenous signals and external context.}
A growing body of work recognises that engagement spikes are often driven
by events external to the platform.
Iamnitchi et al.~\cite{iamnitchi2023modeling} showed that while in-platform signals capture baseline activity, they fail to anticipate the sudden engagement surges tied to external events, which is a gap that exogenous features can fill.
El-Amrany et al.~\cite{elamrany2025viralreddit} introduced the Reddit-V
benchmark of 25{,}000 posts across five subreddit categories and evaluated
several large language models for virality prediction, concluding that
task-specific supervised learning outperforms zero-shot prompting.
Their labelling strategy (top~20\% within each subreddit) is related to
ours, though we adopt a stricter 90th-percentile threshold with an
absolute-score floor and additionally incorporate temporal trend signals
that prior Reddit-focused studies leave unexplored.

\paragraph{Wikipedia pageviews as an attention proxy.}
Wikipedia pageview data has been shown to be a reliable real-time proxy
for public attention.
Yasseri and Bright~\cite{yasseri2016wikipedia} showed that Wikipedia pageview time series can serve as a proxy for collective public attention, demonstrating that surges in views of politically relevant articles track real-world information-seeking behaviour around elections. Wikipedia pageview counts are particularly attractive as an attention signal because they are publicly available at hourly granularity, continuously recorded since 2007, and free of the algorithmic ranking and personalisation effects that confound platform-internal trend signals.

Mesty\'{a}n et al.~\cite{mestyan2013early} exploited Wikipedia activity
to predict movie box-office success, establishing a precedent for using
encyclopedia traffic as a leading indicator of real-world interest.
We extend this line of work by encoding Wikipedia spike terms as
dense embeddings and integrating them via cross-attention, allowing the
model to learn which trending topics are semantically relevant to each
post rather than relying on hand-crafted alignment features.

\paragraph{Cross-attention and multi-stream fusion.}
Cross-attention mechanisms have become a standard tool for
fusing heterogeneous information streams.
In vision--language models, cross-attention conditions one modality on
another to capture fine-grained alignment~\cite{lu2019vilbert}.
Our architecture applies the same principle to a novel pairing:
post-level representations query a temporal trends matrix, enabling the
model to selectively attend to the subset of recent public-interest
topics most aligned with the post's content.

% ── 3. Methodology ────────────────────────────────────────────────────────────
\section{Methodology}

\label{sec:method}

\subsection{Model Overview}

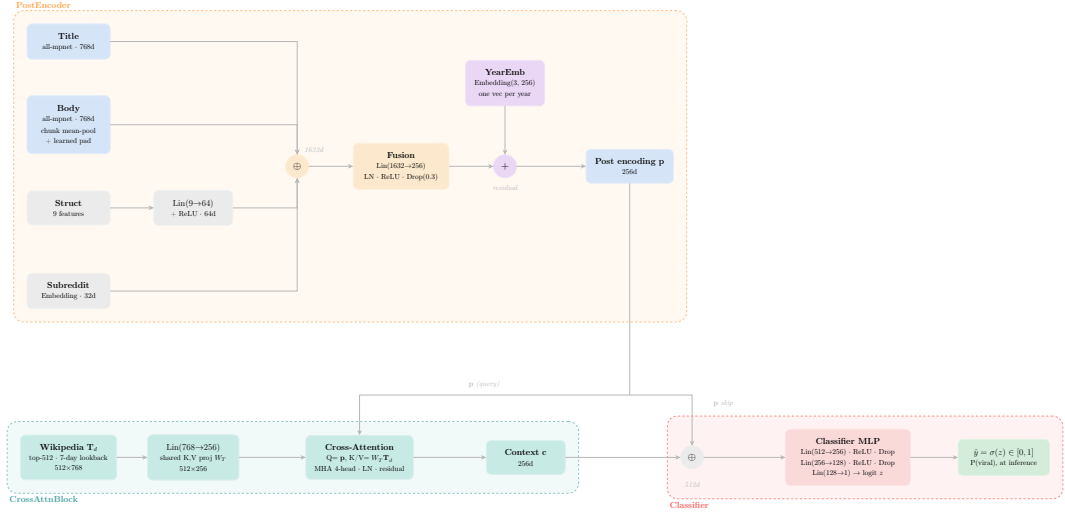
\begin{figure}[H]
\centering
\resizebox{\textwidth}{!}{%
\begin{tikzpicture}

\tikzset{
  B/.style={rectangle,rounded corners=7pt,draw=#1!70,fill=#1,
            align=center,font=\large,inner sep=12pt,minimum height=1.2cm},
  C/.style={circle,draw=#1!80,fill=#1,font=\Large\bfseries,
            minimum size=1.1cm,inner sep=0pt,align=center},
  AR/.style={-{Stealth[length=7pt,width=6pt]},line width=1.4pt,gray!55},
  BR/.style={line width=1.4pt,gray!55},
  DL/.style={font=\normalsize\itshape,text=gray!55},
}

\definecolor{cb}{HTML}{D6E4F7}
\definecolor{co}{HTML}{FCE8CC}
\definecolor{ct}{HTML}{C8EAE4}
\definecolor{cr}{HTML}{FAD9D9}
\definecolor{cv}{HTML}{EAD6F5}
\definecolor{cg}{HTML}{EBEBEB}
\definecolor{cgn}{HTML}{D4EDDA}

% ══ ROW 1: PostEncoder ══════════════════════════════════════════

\node[B=cb,minimum width=4.0cm] (title) at (0, 6)
  {\textbf{Title}\\{\normalsize all-mpnet $\cdot$ 768d}};

\node[B=cb,minimum width=4.0cm] (body) at (0, 2)
  {\textbf{Body}\\{\normalsize all-mpnet $\cdot$ 768d}\\[3pt]
   {\normalsize chunk mean-pool}\\{\normalsize + learned pad}};

\node[B=cg,minimum width=4.0cm] (struct) at (0, -2)                     
  {\textbf{Struct}\\{\normalsize 9 features}};
\node[B=cg,minimum width=3.8cm] (sproj) at (6, -2)
  {Lin(9$\to$64)\\{\normalsize + ReLU $\cdot$ 64d}};                      
\draw[AR] (struct.east) -- (sproj.west);

\node[B=cg,minimum width=4.0cm] (subred) at (0, -6)
  {\textbf{Subreddit}\\{\normalsize Embedding $\cdot$ 32d}};

\node[C=co] (cat1) at (11, 0) {$\oplus$};
\node[DL] at (11.8, 0.8) {1632d}; 

\draw[AR] (title.east) -| (cat1.north);
\draw[AR] (body.east) -| (cat1.north);
\draw[AR] (sproj.east) -| (cat1.south);
\draw[AR] (subred.east) -| (cat1.south);

\node[B=co,minimum width=4.6cm] (fusion) at (16, 0)
  {\textbf{Fusion}\\{\normalsize Lin(1632$\to$256)}\\{\normalsize LN $\cdot$ ReLU $\cdot$ Drop(0.3)}};
\draw[AR] (cat1.east) -- (fusion.west);

\node[C=cv] (addyr) at (21, 0) {$+$};
\node[DL] at (21, -1.0) {residual};
\node[B=cv,minimum width=3.8cm] (yearemb) at (21, 4)
  {\textbf{YearEmb}\\{\normalsize Embedding(3, 256)}\\{\normalsize one vec per year}};
\draw[AR] (fusion.east)   -- (addyr.west);
\draw[AR] (yearemb.south) -- (addyr.north);

\node[B=cb,minimum width=4.2cm] (penc) at (27, 0)
  {\textbf{Post encoding} $\mathbf{p}$\\{\normalsize 256d}};
\draw[AR] (addyr.east) -- (penc.west);

% ══ ROW 2 LEFT: CrossAttnBlock ══════════════════════════════════

\node[B=ct,minimum width=4.4cm] (trends) at (0, -14)
{\textbf{Wikipedia} $\mathbf{T}_d$\\{\normalsize top-512 $\cdot$ 7-day lookback}\\{\normalsize 512$\times$768}};
\node[B=ct,minimum width=4.4cm] (tproj) at (6, -14)
  {Lin(768$\to$256)\\{\normalsize shared K,V proj $W_T$}\\{\normalsize 512$\times$256}};
\draw[AR] (trends.east) -- (tproj.west);

\node[B=ct,minimum width=5.0cm] (xattn) at (14, -14)
  {\textbf{Cross-Attention}\\{\normalsize Q$=\mathbf{p}$, K/V$=W_T\mathbf{T}_d$}\\{\normalsize MHA 4-head $\cdot$ LN $\cdot$ residual}};
\draw[AR] (tproj.east) -- (xattn.west);

\node[B=ct,minimum width=3.8cm] (cenc) at (22, -14)
  {\textbf{Context} $\mathbf{c}$\\{\normalsize 256d}};
\draw[AR] (xattn.east) -- (cenc.west);

% ══ ROW 2 RIGHT: Classifier ═════════════════════════════════════

\node[C=cg] (cat2) at (30, -14) {$\oplus$};
\node[DL] at (30, -15.3) {512d};
\draw[AR] (cenc.east) -- (cat2.west);

\node[B=cr,minimum width=6.0cm] (clf) at (37.5, -14)
  {\textbf{Classifier MLP}\\{\normalsize Lin(512$\to$256) $\cdot$ ReLU $\cdot$ Drop}\\{\normalsize Lin(256$\to$128) $\cdot$ ReLU $\cdot$ Drop}\\{\normalsize Lin(128$\to$1) $\to$ logit $z$}};
\draw[AR] (cat2.east) -- (clf.west);

\node[B=cgn,minimum width=4.4cm] (out) at (45, -14)
  {$\hat{y}=\sigma(z)\in[0,1]$\\{\normalsize P(viral), at inference}};
\draw[AR] (clf.east) -- (out.west);

% ── p cable: shared manifold at y=-11, split to xattn (Q) and cat2 (skip) ──
\coordinate (psplit) at (27, -11);
\draw[BR] (penc.south) -- (psplit);
\draw[AR] (psplit) -- (14, -11) -- (xattn.north);
\node[DL] at (20, -10.5) {$\mathbf{p}$ (query)};
\draw[AR] (psplit) -- (30, -11) -- (cat2.north);
\node[DL] at (31.5, -11.4) {$\mathbf{p}$ skip};

% ══ BACKGROUND ══════════════════════════════════════════════════
\begin{scope}[on background layer]
  \node[draw=orange!50,fill=orange!5,dashed,rounded corners=14pt,
        fit=(title)(body)(struct)(sproj)(subred)(cat1)(fusion)(addyr)(yearemb)(penc),
        inner sep=18pt] (pebox) {};
  \node[anchor=south west,font=\large\bfseries,text=orange!75]
    at (pebox.north west) {PostEncoder};

  \node[draw=teal!45,fill=teal!5,dashed,rounded corners=14pt,
        fit=(trends)(tproj)(xattn)(cenc),
        inner sep=18pt] (cabox) {};
  \node[anchor=north west,font=\large\bfseries,text=teal!75]
    at (cabox.south west) {CrossAttnBlock};

  \node[draw=red!45,fill=red!5,dashed,rounded corners=14pt,
        fit=(cat2)(clf)(out),
        inner sep=18pt] (clfbox) {};
  \node[anchor=north west,font=\large\bfseries,text=red!65]
    at (clfbox.south west) {Classifier};
\end{scope}

\end{tikzpicture}%
}
\caption{\model{} architecture. \textbf{PostEncoder:} Title, Body,
Struct, and Subreddit embeddings are concatenated (1632d), projected
to 256d, and fused with a residual year embedding to form the post
encoding $\mathbf{p}$ (256d). \textbf{CrossAttnBlock + Classifier:}
$\mathbf{p}$ queries the sliding 7-day Wikipedia trends matrix
$\mathbf{T}_d$ via cross-attention to produce context $\mathbf{c}$
(256d). The Classifier maps the concatenation
$[\mathbf{p};\mathbf{c}]$ (512d) through a two-hidden-layer MLP to the
viral logit $z$; $\hat{y} = \sigma(z)$ is applied at inference.}
\label{fig:architecture}
\end{figure}

\subsection{Post Encoder}
The PostEncoder fuses four parallel streams into a single 256-dimensional
post representation $\mathbf{p}$.
\begin{itemize}
    \item \textbf{Title:}
    Post titles are encoded with \texttt{all-mpnet-base-v2}, an MPNet architecture
    \cite{song2020mpnet} fine-tuned using the Sentence-Transformers framework
    \cite{reimers2019sentencebert}, with L2 normalisation, producing
    $\mathbf{t} \in \mathbb{R}^{768}$.
    \item \textbf{Body:}
    The selftext is split into non-overlapping 400-character chunks, each
    encoded with the same model (L2-normalised); the mean of the chunk
    embeddings is L2-renormalised to produce
    $\mathbf{b} \in \mathbb{R}^{768}$. Posts with no body text enter the
    encoder as a zero vector and are substituted at forward time with a
    \emph{learned} padding embedding
    ($\mathbf{b}_{\text{pad}} \in \mathbb{R}^{768}$, trained end-to-end),
    letting the model distinguish a post with no body from one whose body
    carries semantic content. While mean-pooling independent chunks is computationally efficient, it trades off the ability to capture document-level narrative arcs, a limitation we leave for future work using long-context 
encoders.
    \item \textbf{Structure Features:} We pass in 9 structural features per
    post: $\sin(2\pi h/24)$ and $\cos(2\pi h/24)$ for posting hour
    $h$ (so 23:00 and 00:00 sit adjacent), day of the week, log-normalised
    post age, log-normalised title and body lengths, \texttt{has\_body}
    flag, log-normalised subreddit subscriber count, and
    \texttt{is\_weekend} flag. All features are rescaled to $[0,1]$ and
    projected via Linear$(9, 64)$+ReLU to $\mathbf{s} \in \mathbb{R}^{64}$.
    \item \textbf{Subreddit:} Each post's subreddit is mapped to a learned
    $\mathbf{u} \in \mathbb{R}^{32}$ via an
    $\mathrm{Embedding}(N_{\text{sub}}, 32)$ table, trained end-to-end.
    This gives the model a subreddit-specific prior over the fused
    representation without requiring hand-crafted per-community features.
\end{itemize}

\paragraph{Fusion.}
The four streams are concatenated and fused:
\begin{equation}
  \mathbf{p} = \mathbf{e}_y + \mathrm{Drop}\!\left(
    \mathrm{ReLU}\!\left(
      \mathrm{LN}\!\left(
        W_f[\mathbf{t};\mathbf{b};\mathbf{s};\mathbf{u}] + \mathbf{b}_f
      \right)
    \right)
  \right), \quad
  W_f \in \mathbb{R}^{256 \times 1632},
  \label{eq:fusion}
\end{equation}
where $\mathbf{e}_y \in \mathbb{R}^{256}$ is a learned year embedding
added residually so the model can account for the year-on-year shift in
engagement norms.

\subsection{Cross-Attention over Wikipedia Trends}
\paragraph{Trends matrix.}
For each calendar day $d$ we build
$\mathbf{T}_d \in \mathbb{R}^{512 \times 768}$ over a sliding 7-day
lookback window $(d-7,\,d)$ (strictly prior to the post date, so no
future information leaks). Within the window, terms are aggregated by
summing their daily spike popularity; the top 512 are retained, encoded
once with \texttt{all-mpnet-base-v2}, and stacked row-wise. Days with
fewer than 512 window terms are zero-padded, and a key-padding mask
prevents attention from attending to padded rows. If a day has
\emph{no} trending terms at all (every row zero), the attention branch
is skipped and only the post encoding flows through the subsequent
LayerNorm, which avoids a degenerate softmax over an empty key set.

\paragraph{Attention.}
The trends matrix is first projected to the attention dimension via a
single shared linear map, and the post encoding $\mathbf{p}$ is treated as
a single-token query:
\begin{equation}
  \mathbf{c} = \mathrm{LayerNorm}\!\left(
    \mathbf{p} +
    \mathrm{MHA}\!\left(
      \mathbf{p},\;
      W_T \mathbf{T}_d,\;
      W_T \mathbf{T}_d
    \right)
  \right),
  \quad W_T \in \mathbb{R}^{256 \times 768},
  \label{eq:crossattn}
\end{equation}
where $\mathrm{MHA}$ denotes 4-head scaled dot-product attention
($d_k = 64$ per head) with its own internal query/key/value projections,
dropout 0.3, and a key-padding mask derived from zero rows of
$\mathbf{T}_d$. The output $\mathbf{c} \in \mathbb{R}^{256}$ captures
which recent trending topics are most semantically aligned with the post.

\subsection{Classifier and Training}
\paragraph{Classifier.}
The concatenation $[\mathbf{p};\mathbf{c}] \in \mathbb{R}^{512}$ is passed
through Linear$(512, 256)$--ReLU--Dropout$(0.3)$--Linear$(256, 128)$--
ReLU--Dropout$(0.3)$--Linear$(128, 1)$ to produce a viral logit $z$.
Probabilities $\hat{y} = \sigma(z) \in [0,1]$ are obtained by applying
the sigmoid at inference.

\paragraph{Training.}
We minimise focal loss \cite{lin2017focal} with $\alpha=0.25$ and
$\gamma=2.0$, which down-weights easy negatives and focuses learning on
hard-to-classify examples---well suited for our
9.7\%/90.3\% class imbalance. Optimisation uses AdamW
($\mathrm{lr}=10^{-4}$, $\lambda=10^{-5}$) with a linear warmup over the
first 2 epochs (10\%$\to$100\% of base LR) followed by cosine annealing
over the remaining 23 epochs. We train for up to 25 epochs with batch
size 64, gradient clipping at $\lVert g \rVert_2 \le 1$, and early
stopping on validation AUC-PR (patience 5). Full hyperparameters are
listed in Appendix~\ref{app:hyperparams}.
% ── 5. Experiments ────────────────────────────────────────────────────────────
\section{Experiments}
\label{sec:experiments}

\subsection{Data}
\label{sec:data}

\subsubsection{Reddit Posts}

We collected Reddit posts via the Arctic Shift API
(\url{https://arctic-shift.photon-reddit.com}), which provides access to
a public archive of Reddit submissions. We selected 10 subreddits
spanning news, technology, finance, science, and entertainment:
\textbf{r/worldnews}, \textbf{r/politics}, \textbf{r/technology},
\textbf{r/stocks}, \textbf{r/wallstreetbets}, \textbf{r/science},
\textbf{r/futurology}, \textbf{r/movies}, \textbf{r/television},
and \textbf{r/gaming}. This mix was chosen to balance high-volume
discussion communities with more text-heavy ones and to cover distinct
topical regimes so that the Wikipedia trend signal is tested against
different content distributions rather than a single editorial tone.
We sampled up to 1{,}000 posts per 7-day window per subreddit from
January 2021 through December 2023 (3 years), yielding
788{,}873 posts after removing deleted
(\texttt{[deleted]}) and moderated (\texttt{[removed]}) submissions.

\subsubsection{Virality Labels}

Virality is defined per-subreddit to control for community-specific
engagement scales. For each post we compute an engagement score
$v= \text{score} + 0.3 \cdot \text{num\_comments}$, and label a post
viral if it exceeds the 90th percentile of $v$ within its subreddit and
its raw score is at least 100. The absolute-score floor discards posts
that beat a small subreddit's percentile threshold despite having
negligible engagement. The resulting positive rate is
9.7\%, giving a class-imbalance ratio of
$n_{-}/n_{+} \approx 9.3$, which motivates our use of focal loss
(Section~\ref{sec:method}).

\subsubsection{Wikipedia Trend Signals}

As a proxy for real-world public attention we processed English
Wikipedia pageview data from the public BigQuery dataset
(\texttt{bigquery-public-data.wikipedia.pageviews\_\{year\}}). Spike
detection operates at \emph{daily} granularity, not weekly.

For each article $i$ and day $d$, let $v_{i,d}$ denote total English
pageview count on day $d$. We define the \textbf{spike ratio} and
\textbf{absolute spike} as
\begin{equation}
  \rho_{i,d} = \frac{v_{i,d}}{v_{i,d-1}}, \qquad
  \delta_{i,d} = v_{i,d} - v_{i,d-1},
  \label{eq:spike}
\end{equation}
with an article's first appearance ($v_{i,d-1}$ undefined) treated as
an implicit spike. We exclude disambiguation-style titles
(\texttt{'\%:\%'}), the \texttt{Main\_Page}, and a small set of
scraping-artifact titles, and require $v_{i,d} > 3{,}000$ to filter
low-traffic noise. An article is considered \textbf{trending} on day
$d$ if
\begin{equation}
  \rho_{i,d} \geq 2.0,
  \label{eq:spike_filter}
\end{equation}
and we rank candidates by a composite spike score
$\rho_{i,d} \cdot \log(\delta_{i,d} + 1)$, retaining the top-512 per
day.
A post-hoc filtering stage further removes known evergreen and
navigational pages (\eg\ \textit{Google Classroom}, \textit{Microsoft
Teams}) via an explicit blocklist, regex patterns for structural noise,
and an auto-evergreen detector that drops terms appearing in more than
60\% of all days.
Ranking by the composite score surfaces high-traffic events (\eg\ a
celebrity death with $+$400{,}000 views) over low-traffic outliers
whose ratios are arbitrarily large (\eg\ an obscure article jumping
from 10 to 100 views).

Trend matrices $\mathbf{T}_d$ used by the cross-attention block
(Section~\ref{sec:method}) are then constructed per day by aggregating
these daily trending terms over the preceding 7-day window $(d-7, d)$,
ranking by summed popularity score, taking the top 512, and encoding
each title once with \texttt{all-mpnet-base-v2}.

\subsection{Experimental Setup}

We split the dataset 70/15/15 into train/validation/test sets,
stratified by (year, virality label) to preserve both temporal and
class distributions. 
We deliberately use random stratified splitting rather than a forward temporal split because our research question concerns \emph{contemporaneous alignment}---given a post and the public-attention landscape surrounding its posting date, can the model identify whether it will go viral?---rather than forecasting future virality from past trends.
All results are reported on the held-out test set
(118{,}331 posts, 11{,}459
viral). We use AUC-PR as our primary metric, which is appropriate for
severely imbalanced classification, and additionally report AUC-ROC
and F1 at the fixed threshold of 0.31, selected by maximising F1 on the full model's validation set and applied uniformly to all variants.

\subsection{Baselines}
We compare \model{} against:
\begin{itemize}
  \item \textbf{Text Only:} Title, body, structure, and subreddit
    streams only; no year embedding and no cross-attention.
  \item \textbf{Text + Year:} The Text Only encoder with the learned
    year embedding added residually.
  \item \textbf{Text + Trends:} The Text Only encoder with cross-attention
    over $\mathbf{T}_d$ but no year embedding.
\end{itemize}

\section{Results}
\label{sec:results}

\subsection{Main Results}

\begin{table}[h]
\centering
\caption{Test-set performance of \model{} and ablation variants.
AUC-PR is the primary metric; $\Delta$ denotes improvement over the
text-only baseline. F1 is reported at the fixed threshold of 0.31.}
\label{tab:results}
\begin{tabular}{lcccc}
\toprule
Model & AUC-PR & $\Delta$ & AUC-ROC & F1 @ $t^{*}{=}0.31$ \\
\midrule
Text Only              & 0.3431 & ---      & 0.8275 & 0.3869 \\
Text + Year            & 0.3554 & $+$0.012 & 0.8338 & 0.3952 \\
Text + Trends          & 0.3476 & $+$0.005 & 0.8316 & 0.3452 \\
\model{} (Full)        & \textbf{0.3584} & $+$0.015 & \textbf{0.8355} & \textbf{0.3986} \\
\bottomrule
\end{tabular}
\end{table}

\begin{figure}[h]
\centering
\begin{subfigure}[t]{0.48\textwidth}
  \centering
  \includegraphics[width=\linewidth]{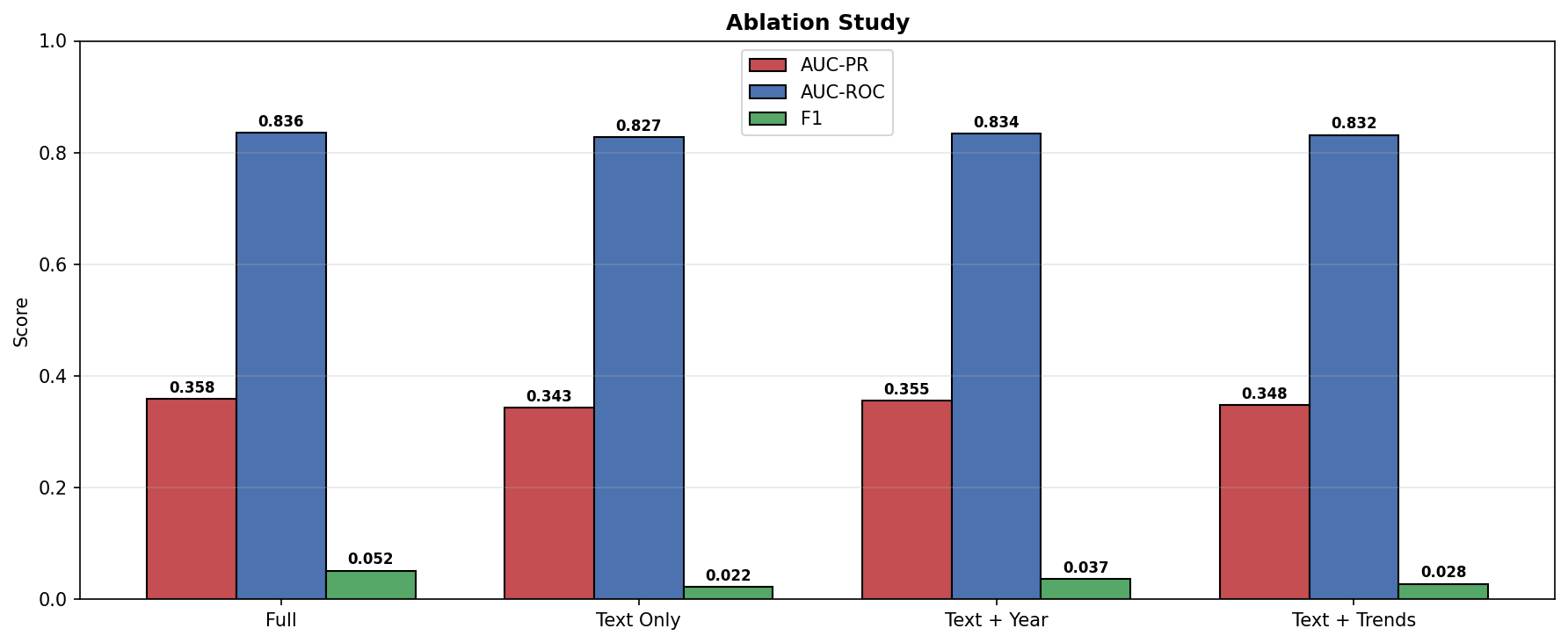}
  \caption{Ablation study: AUC-PR, AUC-ROC, and F1 across all four
  model variants.}
  \label{fig:ablation}
\end{subfigure}
\hfill
\begin{subfigure}[t]{0.48\textwidth}
  \centering
  \includegraphics[width=\linewidth]{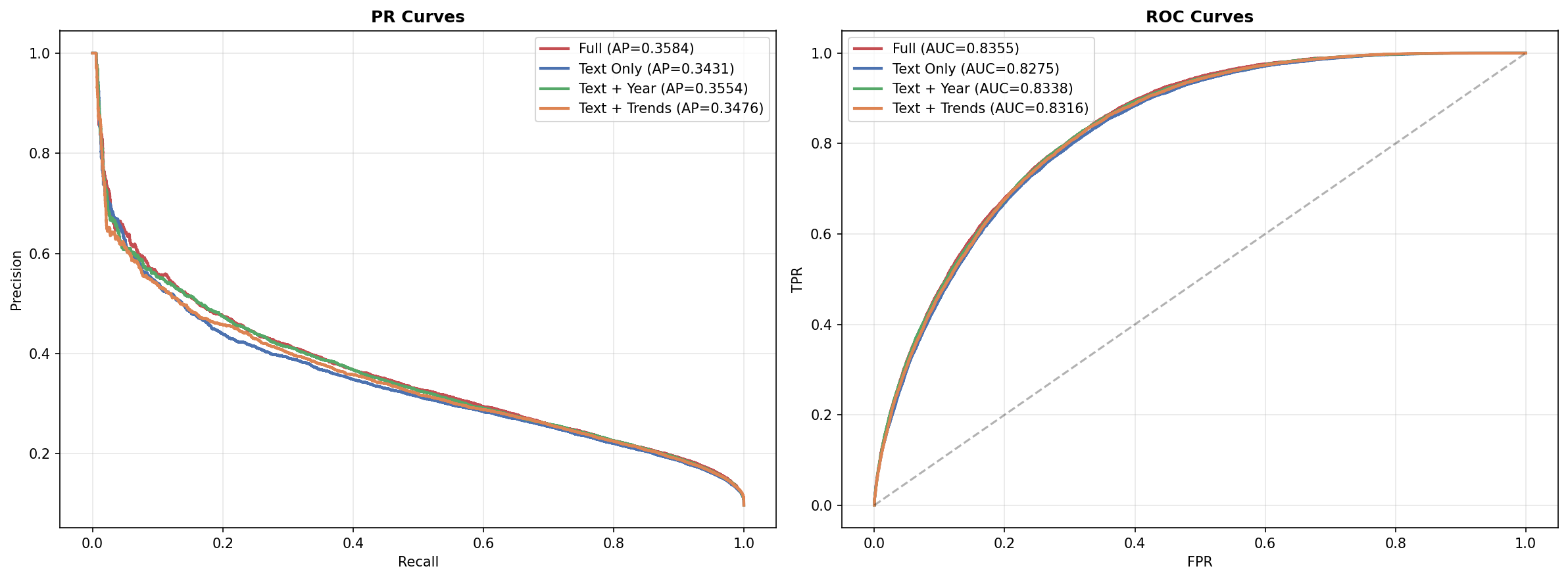}
  \caption{Precision--Recall and ROC curves for all ablation variants.}
  \label{fig:pr_roc}
\end{subfigure}
\caption{Model evaluation across ablation variants.
(a)~The full model achieves the highest scores across all three metrics.
(b)~PR and ROC curves show consistent separation, with the full model
dominating at most operating points.}
\label{fig:results}
\end{figure}

The full model's advantage over the text-only baseline ($+$0.015 AUC-PR)
is modest but consistent across metrics
(Table~\ref{tab:results}, Figure~\ref{fig:results}).
Notably, the year embedding contributes substantially more than trend cross-attention. Adding year embeddings alone to the text-only baseline yields $+$0.012 AUC-PR, while adding trend cross-attention on top yields a further $+$0.003. We attribute the year embedding's contribution to the substantial shift in per-year engagement norms across 2021--2023, and treat the trend cross-attention's contribution as a smaller, complementary signal that nonetheless drives the full model to the highest AUC-ROC (0.836). The strong AUC-ROC (
$>$0.83) across all variants suggests the learned post-text features capture genuine virality signal.

Due to the severity of class imbalance (9.7\% positive rate), the default classification threshold of 0.5 yields artificially low recall. Consequently, we report F1 at the fixed threshold of 0.31, which substantially improves detection of viral posts and provides a more accurate reflection of the model's practical utility (Section~\ref{sec:threshold}). Interestingly, adding trend signals without the year embedding (Text + Trends) decreases the F1 score relative to the baseline, suggesting that exogenous trends require temporal grounding to be utilized effectively.

\subsection{Statistical Significance}
\label{sec:bootstrap}

To assess the reliability of the observed differences, we compute 95\%
bootstrap confidence intervals (1{,}000 resamples with replacement from
the test set) for each metric across all four model variants.

\begin{figure}[h]
\centering
\includegraphics[width=\textwidth]{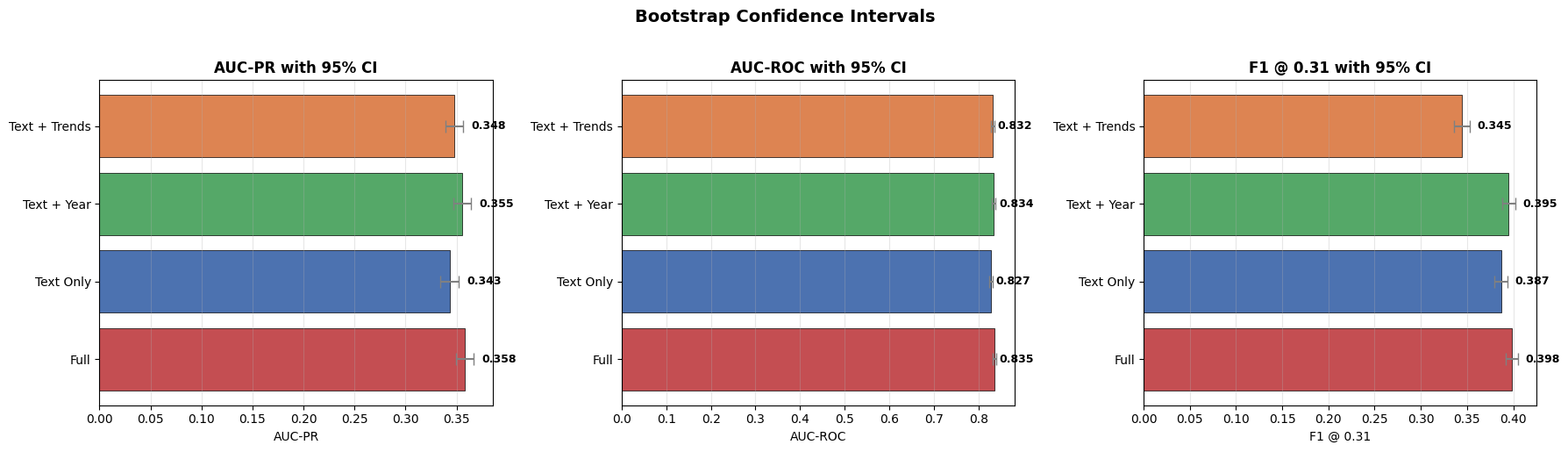}
\caption{Bootstrap 95\% confidence intervals for AUC-PR, AUC-ROC, and F1
across all ablation variants (1{,}000 resamples).  The full model
consistently achieves the highest point estimates, though intervals overlap
with Text$+$Year on AUC-PR, suggesting that the trend signal provides a modest contribution.}
\label{fig:bootstrap}
\end{figure}

Figure~\ref{fig:bootstrap} confirms that the full model achieves the highest
point estimates across all metrics.  The AUC-PR confidence intervals for the
full model and Text$+$Year partially overlap, consistent with the modest
$+$0.003 gap between these variants.  However, both clearly separate from the
text-only baseline, validating the contributions of temporal and trend signals.
The F1 intervals show wider spread due to the threshold sensitivity inherent
in highly imbalanced classification.

\subsection{Per-Subreddit Performance}
\label{sec:persubr}

\begin{figure}[h]
\centering
\includegraphics[width=\textwidth]{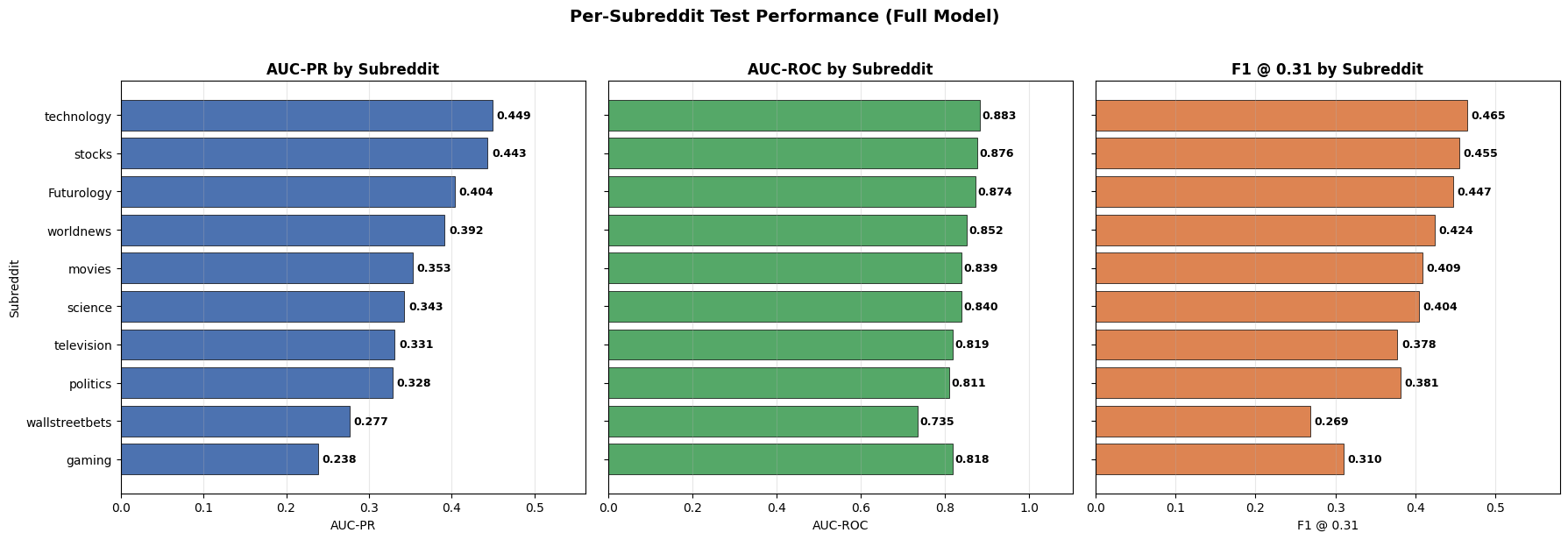}
\caption{Per-subreddit test-set performance of the full model (AUC-PR,
AUC-ROC, F1).  Performance varies substantially across communities, with
r/technology (0.449) and r/stocks (0.443) achieving the highest AUC-PR,
while r/gaming (0.238) proves most challenging.}
\label{fig:persubr}
\end{figure}

Figure~\ref{fig:persubr} disaggregates the full model's performance by
subreddit.  AUC-PR ranges from 0.238 (r/gaming) to 0.449
(r/technology), a nearly $2\times$ spread that reflects fundamental
differences in what drives virality across communities.  Subreddits with
strong topical overlap with Wikipedia trend content---such as
r/technology, r/stocks, r/futurology, and r/worldnews---achieve the
highest AUC-PR ($>$0.39), consistent with our hypothesis that exogenous
attention signals benefit content whose virality is partly driven by
external events.  Conversely, r/gaming and r/wallstreetbets, where
community-specific culture dominates engagement, prove most challenging.
AUC-ROC remains high across all communities ($\geq$0.735), confirming
that the model captures discriminative signal even where precision is
limited by class imbalance.

\subsection{Threshold Optimisation}
\label{sec:threshold}

\begin{figure}[h]
\centering
\includegraphics[width=\textwidth]{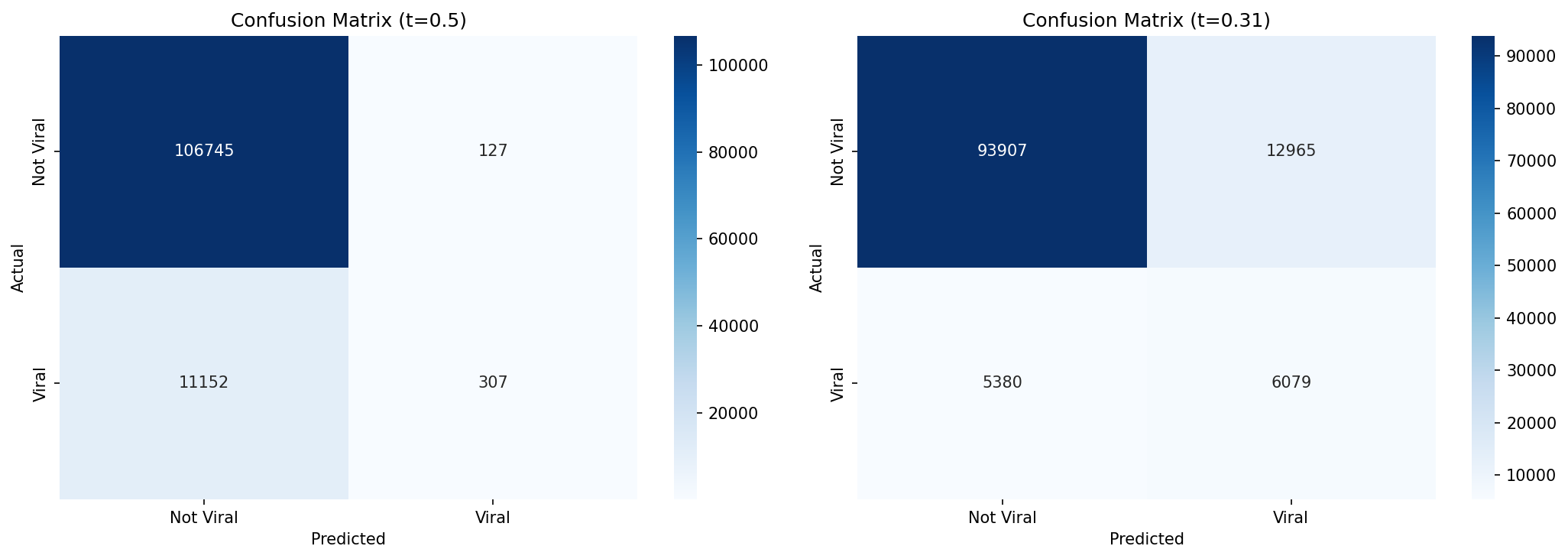}
\caption{Confusion matrices at the default threshold ($t=0.5$, left) and
the validation-optimised threshold ($t=0.31$, right).  Lowering the
threshold increases true positive detections from 307 to 6{,}079
(recall from 2.7\% to 53.0\%), at the cost of additional false positives
(127 to 12{,}965).}
\label{fig:confusion}
\end{figure}

The default classification threshold of 0.5 yields low recall (2.7\%)
due to the model's conservative probability estimates under severe class
imbalance (Figure~\ref{fig:confusion}, left).  Sweeping the threshold on
the validation set identifies $t^{*} = 0.31$ as optimal for F1,
improving test-set F1 from 0.052 to approximately 0.40
(Figure~\ref{fig:confusion}, right).  At $t^{*} = 0.31$, the model
correctly identifies 6{,}079 of 11{,}459 viral posts (53.0\% recall)
with precision of 31.9\% (6{,}079 out of 19{,}044 predicted positives).
This precision--recall trade-off is consistent with the practical use
case of surfacing candidate viral posts for further review, where
moderate false positive rates are acceptable.

\subsection{Qualitative Analysis}

Inspection of misclassified examples reveals interpretable patterns.
False positives tend to be posts whose titles closely align with
trending Wikipedia topics in embedding space. For instance, a
low-engagement r/technology post about a newsworthy company
receiving high predicted probability because its title was semantically
similar to that week's trending terms.
Conversely, the hardest false negatives are posts that achieved
virality through community-specific humour, personal relatability, or
outrage---content whose viral potential derives from social dynamics
rather than topical alignment with external trends.

% ── 7. Limitations ────────────────────────────────────────────────────────────
\section{Limitations and Future Work}
\label{sec:limitations}

\paragraph{Data scope.}
Our dataset spans 10 subreddits over 3 years.
Reddit communities are topically siloed, and filter-bubble effects within
subreddits may skew engagement patterns in ways that limit
generalisability.
Expanding to a broader set of subreddits, additional years, and
cross-platform data would strengthen external validity.

\paragraph{Static engagement scores.}
Arctic Shift provides final engagement counts at archival time, not early
engagement velocity.
Early engagement is a known strong predictor of eventual virality~\cite{szabo2010predicting}; incorporating streaming signals remains future
work.

\paragraph{Heuristic design choices.}
Several pipeline components rely on heuristic decisions, including the
comment weight $\beta$ in the engagement score and the 400-character body
truncation length.
While these choices are validated through sensitivity analysis
(Appendix~\ref{app:sensitivity}), a more principled approach with learned weights from end to end could improve model performance.

\paragraph{Trends proxy.}
Wikipedia pageviews are a reasonable but imperfect proxy for real-time
public interest.
The Google Trends API now offers daily search-interest data;
integrating it could complement or replace our Wikipedia signal with a
broader measure of attention.
Integration remains future work.

\paragraph{Model scale.}
\model{} uses frozen \texttt{all-mpnet-base-v2} (768d) encodings.
Fine-tuning this encoder on Reddit text end-to-end, or swapping in a
domain-adapted temporal language model, may substantially improve title
and body representations, at the cost of significantly higher training
compute.

\paragraph{Contemporaneous vs.\ forward prediction.}
Our evaluation measures contemporaneous alignment between posts and concurrent attention signals, not forecasting performance. \model{} is not designed to predict virality of posts written after the latest training date, and we make no claim about its forward-temporal generalization. Adapting the architecture for true forecasting---e.g., by using only past trend information at training time and held-out future periods at test time---would be a meaningful extension but addresses a different research question.

\paragraph{Broader impacts.}
While \model{} is designed for observational and analytical purposes, architectures 
that accurately predict virality using temporal alignment could theoretically be 
repurposed by bad actors to optimise the spread of polarised content, spam, or 
computational propaganda. We encourage future work to explore how platforms can 
use such predictive signals defensively to mitigate coordinated engagement 
manipulation.

% ── 8. Conclusion ─────────────────────────────────────────────────────────────
\section{Conclusion}
\label{sec:conclusion}

We presented \model{}, a virality prediction model that fuses four
post-level streams with a cross-attention block over daily sliding-window
Wikipedia trend signals. Our central finding is that posts whose content
aligns semantically with contemporaneous spikes in public attention are
measurably more likely to go viral, and that this alignment can be
captured by attending over a compact daily trends matrix rather than
relying on hand-crafted features. The full model achieves AUC-PR 0.3584
and AUC-ROC 0.8355 on a held-out test set of 118{,}331 Reddit posts,
with ablations indicating additive contributions from both temporal
year embeddings ($+$0.012 AUC-PR) and trend cross-attention
($+$0.003 AUC-PR over Text+Year). Per-subreddit analysis further shows
that the trend signal helps most where it should: in topically
externally-driven communities such as r/technology, r/stocks, and
r/worldnews, while community-culture-driven subreddits like r/gaming
remain harder to predict from external signals alone.

Taken together, these results offer evidence that exogenous attention
signals---even when approximated by a free, publicly-archived source
like Wikipedia pageviews---constitute a useful complement to in-platform
features for content virality prediction. We see two natural extensions.
First, integrating richer attention proxies (Google Trends, news-volume
data, cross-platform engagement) may sharpen the trend representation
beyond what Wikipedia alone provides. Second, replacing the frozen
encoder with one fine-tuned end-to-end on Reddit text, and incorporating
early engagement velocity rather than only static post-level features,
would address the two limitations most likely to constrain real-world
deployment. More broadly, we view virality prediction not as a fixed
function of post content, but as a problem of aligning content with the
moving target of public attention---a framing in which exogenous
signals are not optional, but essential.

% ── References ────────────────────────────────────────────────────────────────
\bibliographystyle{plainnat}

\appendix

% ── A. Hyperparameters ────────────────────────────────────────────────────────
\section{Hyperparameters and Implementation Details}
\label{app:hyperparams}

\begin{table}[H]
\centering
\caption{Full hyperparameter configuration.}
\begin{tabular}{ll}
\toprule
Hyperparameter & Value \\
\midrule
Sentence encoder & \texttt{all-mpnet-base-v2} (768d) \\
Text embedding dim & 768 \\
Hidden dim & 256 \\
Structured feature dim & 9 \\
Structured hidden dim & 64 \\
Cross-attention heads & 4 \\
Trends sequence length & 512 \\
Dropout & 0.3 \\
Batch size & 64 \\
Optimiser & AdamW ($\text{lr}=10^{-4}$, $\lambda=10^{-5}$) \\
LR schedule & Linear warmup (2 epochs, 10\%$\to$100\%) + Cosine annealing \\
Loss & Focal loss ($\alpha=0.25$, $\gamma=2.0$) \\
Max epochs & 25 \\
Early stopping patience & 5 (val AUC-PR) \\
Virality percentile & 90th (top 10\% = viral) \\
Min viral score & 100 \\
Body chunk size & 400 chars \\
Wikipedia spike threshold & $\geq 2.0\times$ day-over-day ratio \\
Wikipedia terms per day & Top 512 by composite spike score \\
Random seed & 42 \\
\bottomrule
\end{tabular}
\end{table}

\section{Sensitivity Analysis: Comment Weight $\beta$}
\label{app:sensitivity}

We sweep the comment weight $\beta$ in $v = s + \beta \cdot c$ over
$\{0.0, 0.1, 0.2, 0.3, 0.5, 1.0, 2.0\}$ and measure label stability
against the $\beta = 0.3$ baseline.

\begin{table}[H]
\centering
\caption{Label stability across $\beta$ values. Jaccard and flip rate
are computed against the $\beta = 0.3$ baseline labelling.}
\label{tab:beta_sensitivity}
\begin{tabular}{lcccc}
\toprule
$\beta$ & Viral count & Viral \% & Jaccard & \% flipped \\
\midrule
0.0 & 76{,}916 & 9.75 & 0.9731 & 0.26 \\
0.1 & 76{,}744 & 9.73 & 0.9846 & 0.15 \\
0.2 & 76{,}558 & 9.70 & 0.9928 & 0.07 \\
0.3 & 76{,}392 & 9.68 & 1.0000 & --- \\
0.5 & 76{,}086 & 9.64 & 0.9860 & 0.14 \\
1.0 & 75{,}520 & 9.57 & 0.9589 & 0.40 \\
2.0 & 74{,}823 & 9.48 & 0.9230 & 0.77 \\
\bottomrule
\end{tabular}
\end{table}

Across $\beta \in [0.1, 0.5]$, fewer than 0.2\% of labels flip and
Jaccard similarity exceeds 0.98. Even at the extremes, flip rates
remain below 1\%, indicating that the subreddit-stratified
90th-percentile threshold dominates the labelling decision and our
results are not sensitive to the specific comment weight chosen.

\section{Structured Feature Descriptions}
\label{app:features}

\begin{table}[H]
\centering
\caption{The nine structured features used in Stream~3.}
\begin{tabular}{lll}
\toprule
Index & Feature & Normalisation \\
\midrule
0 & $\sin(2\pi h/24)$ & Cyclical \\
1 & $\cos(2\pi h/24)$ & Cyclical \\
2 & Day of week (0=Mon, 6=Sun) & Divided by 6 \\
3 & Post age (days since 2005-06-23) & $\log_{10}$, then max-norm \\
4 & Title character length & $\log_{1p}$, then max-norm \\
5 & Body character length (0 if empty) & $\log_{1p}$, then max-norm \\
6 & Has body (binary) & Binary \\
7 & $\log_{10}(\text{subscribers}+1)$ & Max-norm \\
8 & Is weekend (binary) & Binary \\
\bottomrule
\end{tabular}
\end{table}

\section{Training Convergence}
\label{app:training}

\begin{figure}[H]
\centering
\includegraphics[width=\textwidth]{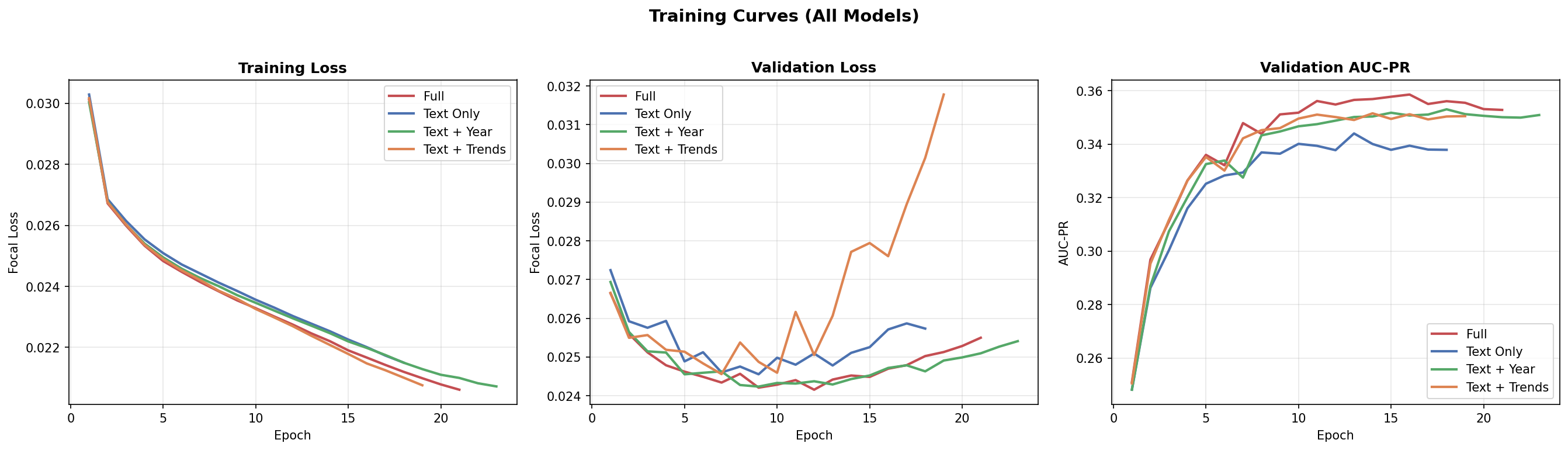}
\caption{Training loss, validation loss, and validation AUC-PR across
epochs for all four model variants.  All models converge within 20 epochs;
validation loss begins to increase after approximately epoch 10 while
AUC-PR plateaus, indicating that early stopping on AUC-PR (patience~5)
correctly prevents overfitting.}
\label{fig:training}
\end{figure}

\section{Threshold Optimisation Sweep}
\label{app:threshold}

\begin{figure}[H]
\centering
\includegraphics[width=0.7\textwidth]{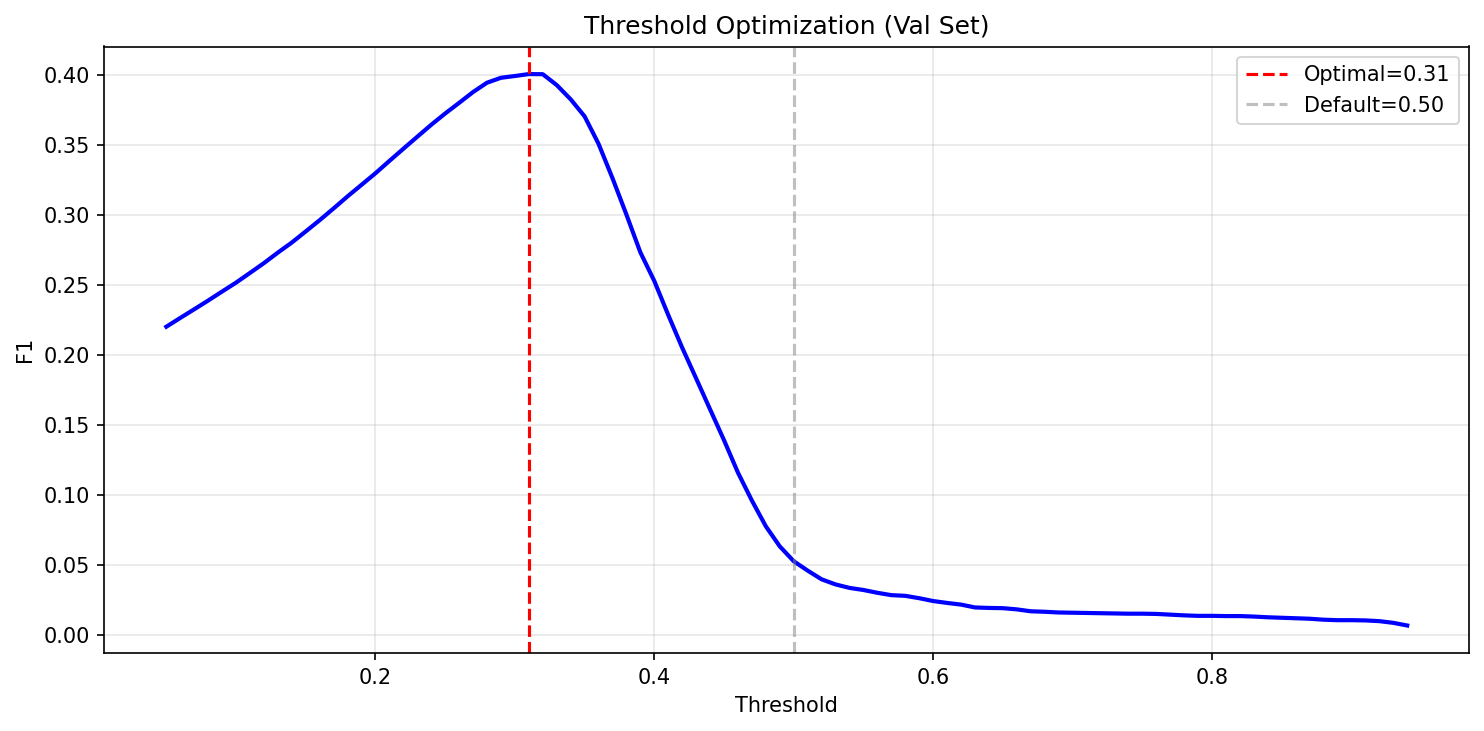}
\caption{F1 score as a function of classification threshold on the
validation set.  The optimal threshold $t^{*}=0.31$ achieves F1$\approx$0.40,
compared to F1$=$0.052 at the default $t=0.5$.}
\label{fig:threshold_sweep}
\end{figure}

\section{Calibration Analysis}
\label{app:calibration}

\begin{figure}[H]
\centering
\includegraphics[width=\textwidth]{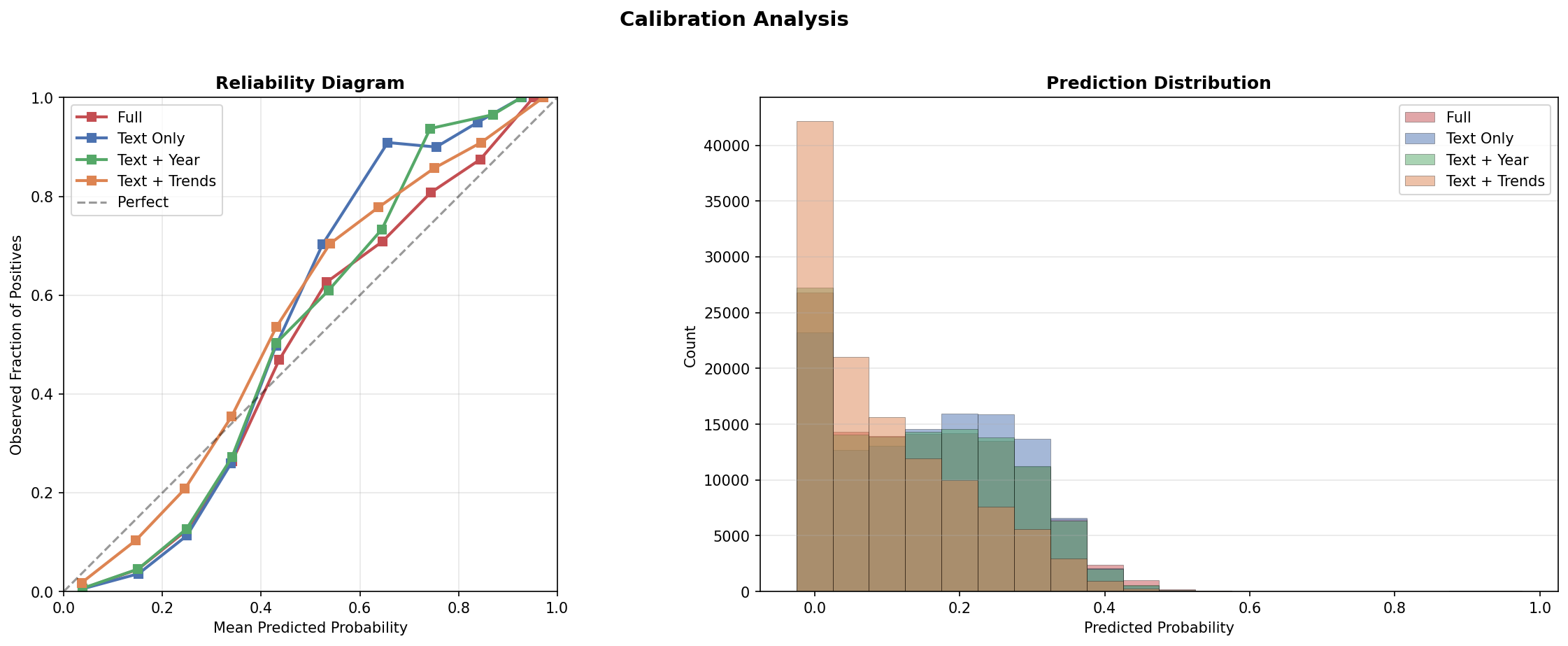}
\caption{Reliability diagram (left) and prediction score distribution
(right) for all four model variants.  All variants show reasonable
calibration up to predicted probabilities of ${\sim}0.5$, with the
majority of predictions concentrated below 0.4 due to the low base rate
(9.7\%).}
\label{fig:calibration}
\end{figure}

\section{Representation Visualisation}
\label{app:tsne}

\begin{figure}[H]
\centering
\includegraphics[width=\textwidth]{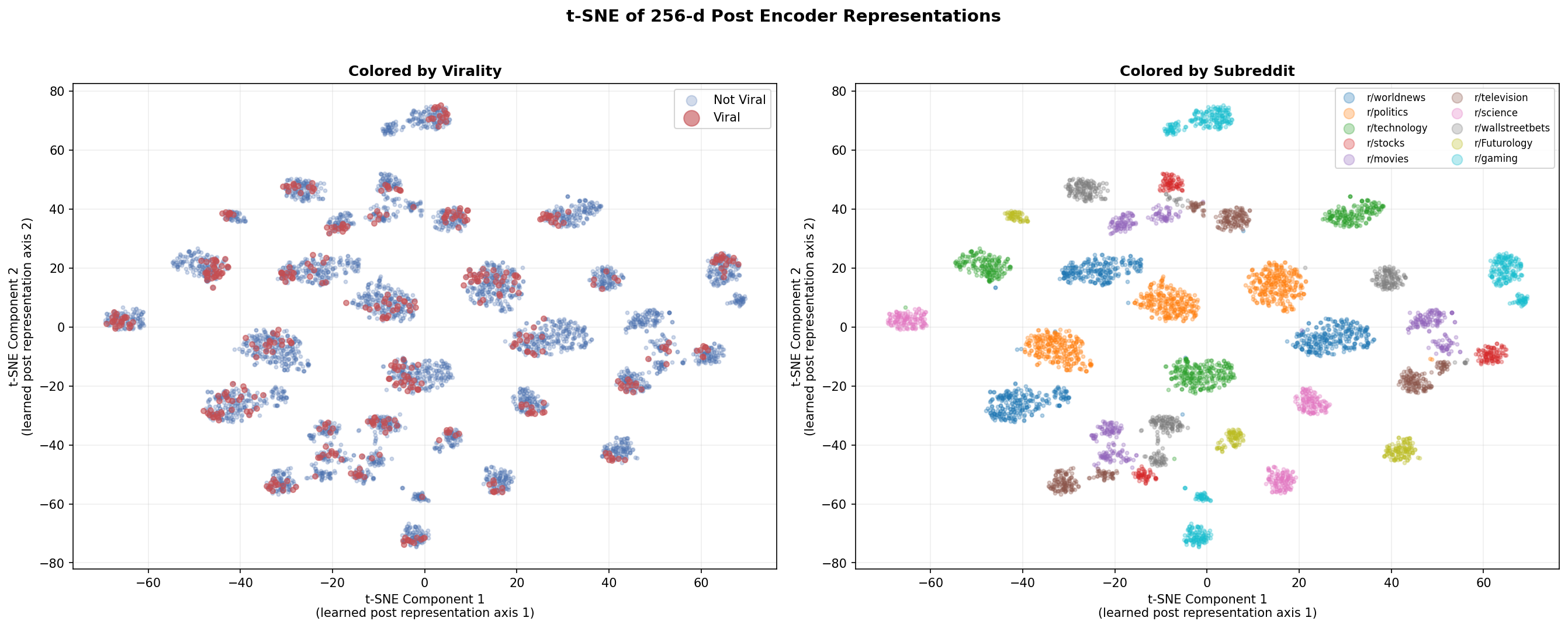}
\caption{t-SNE projection of 256-dimensional PostEncoder representations,
coloured by virality label (left) and subreddit (right).  Subreddit
clusters are clearly separated, confirming that the encoder captures
community-specific features.  Viral posts (red, left) are distributed
across clusters rather than forming a single cluster, consistent with
virality being a within-community phenomenon.}
\label{fig:tsne}
\end{figure}

\section{Reproducibility}
\label{app:reproducibility}
To ensure the reproducibility of our findings, we have made our complete codebase publicly available. The repository contains the data collection and preprocessing scripts for both the Reddit API and BigQuery Wikipedia pageviews, the PyTorch implementation of the \model{} architecture, and the training and evaluation pipelines required to replicate the ablation studies and threshold optimisation discussed in this paper. 

The code, along with instructions for setting up the environment, can be accessed at: \\ \url{https://github.com/som1shi/virality-predictor-model}

\end{document}